\documentclass[letterpaper, 10 pt, conference]{ieeeconf}

\usepackage{graphics} 
\usepackage{subfigure}
\usepackage{epsfig} 
\usepackage{amsmath} 
\usepackage[T1]{fontenc}
\usepackage{bm}
\usepackage{cite}
\usepackage{here}
\usepackage{booktabs}
\usepackage{threeparttable}
\usepackage{multirow}

\IEEEoverridecommandlockouts                              
\title{\LARGE \bf Adaptive Vision-Based Control of Redundant Robots with Null-Space Interaction for Human-Robot Collaboration}
\author{Xiangjie Yan, Chen Chen, and Xiang Li
\thanks{X. Yan, C. Chen, and X. Li are with Department of Automation, Tsinghua University. 
This work was supported in part by the Science and Technology Innovation 2030-Key Project under Grant 2021ZD0201404, in part by Beijing National Research Center for Information Science and Technology, and in part by the National Natural Science Foundation of China under Grant U21A20517.  Corresponding author: Xiang Li (xiangli@tsinghua.edu.cn)}
}

\begin{document}
\maketitle

\begin{abstract}
Human-robot collaboration aims to extend human ability through cooperation with robots. This technology is currently helping people with physical disabilities, has transformed the manufacturing process of companies, improved surgical performance, and will likely revolutionize the daily lives of everyone in the future. Being able to enhance the performance of both sides, such that human-robot collaboration outperforms a single robot/human, remains an open issue.
For safer and more effective collaboration, a new control scheme has been proposed for redundant robots in this paper, consisting of an adaptive vision-based control term in task space and an interactive control term in null space. Such a formulation allows the robot to autonomously carry out tasks in an unknown environment without prior calibration while also interacting with humans to deal with unforeseen changes (e.g., potential collision, temporary needs) under the redundant configuration. The decoupling between task space and null space helps to explore the collaboration safely and effectively without affecting the main task of the robot end-effector.
The stability of the closed-loop system has been rigorously proved with Lyapunov methods, and both the convergence of the position error in task space and that of the damping model in null space are guaranteed.
The experimental results of a robot manipulator guided with the technology of augmented reality (AR) are presented to illustrate the performance of the control scheme.
\end{abstract}

\section{Introduction}
An increasing amount of attention has been given to human-robot collaboration, which combines human cognitive skills, dexterity, and flexibility with robots' advantage to complement tasks requiring high precision, high speed, high repetition and the response to unforeseen changes. Several applications of human-robot collaboration can be found in space/deep-sea exploration \cite{douglas1995real,brantner2021controlling}, surgical/rehabilitation scenarios \cite{surgical_application,casas2020human} and the manufacturing industry \cite{manufacture_application}. Recently, a practical application comes in the form of a nasal swab robot, which can assist with diagnosing COVID-19 \cite{COVID19}.

Human-robot collaboration focuses on effectively combining the contribution of both humans and robots.
A straightforward approach to human-robot collaboration would be to suspend the current main task of the robot end-effector, and then force it to follow human guidance, and finally switch it back to implement the main task automatically after that\cite{TCST16,lyu2019human}. To guarantee efficiency, considerable works have also been done to establish the framework where both human and robot work simultaneously, through direct contact, indirect contact (mediated by a third object), and teleoperation \cite{de2008atlas,losey2018review}. 
A direct-contact formulation is presented in \cite{tro14_sadeghian}, where a null-space impedance controller with observers is proposed for redundant robots, which maintains the compliance of the robot's body during physical contact with humans or its surrounding environment.
Collaboration through indirect contact can be found in various tasks, such as collaboratively carrying a table through a door \cite{ijrr17}, polishing\cite{peternel2018online}, and manipulating a large object \cite{stouraitis2020online}, where the robot continuously adjusts its role to suit the human motion or intention.  
One example of teleoperated collaboration is surgery \cite{icra_2018}, where the surgeon is dominant in controlling the position of the robot’s end-effector, while the robot autonomously varies the end-effector’s orientation according to the task and environment.
In the aforementioned works, the role division is commonly designed beforehand, in the sense that less flexibility may be provided to deal with the unforeseen changes during the task; for example, a suddenly appearing obstacle which may crash into the robot. In such scenarios, the role-division may not be able to respond effectively or even fail. Hence, a more dynamic formulation that allows for human intervention at any time is needed.

For a robot working with humans in an uncalibrated environment (e.g., surgical environment, cluttered industrial workshop), the robot control scheme must be able to deal with various uncertainties. Much progress has been achieved in the robust/adaptive control of robot manipulators. To overcome the problem of uncertain kinematics, Cheah {\em et al.} proposed the adaptive Jacobian robot controller with the online estimation of both the kinematic and dynamic parameters \cite{cheah2006adaptive}. Liu {\em et al.} proposed an uncalibrated vision-based control in eye-to-hand configuration where the depth information was unknown \cite{liu2006uncalibrated}. Liang {\em et al.} extended that to a unified adaptive method for both eye-in-hand and eye-to-hand configuration \cite{liang2015unified}.
An adaptive task space controller without measuring task space or joint-space velocity is introduced in \cite{liang2010adaptive}. 
However, the aforementioned control methods are designed for isolated robots that do not interact with humans. Human participation is likely to disturb the adaptation of unknown parameters and hence deteriorate the control performance. 

Note that most existing frameworks for human-robot collaboration commonly focus on the task of robot end-effector by shaping its position or trajectory. However
for some manipulation tasks, especially in an uncalibrated environment, the robot's body should also be controlled to suit the unstructured surface or cramped space (e.g., a shipwreck salvage or surgical environment), or to avoid the collision of unforeseen obstacles. In these scenarios, the robot may not be able to respond promptly or perceive it due to having a limited sensing zone, and hence utilizing human knowledge or experience to handle such issues would be an effective solution.

In an attempt to fulfill this requirement, this paper presents a new adaptive vision-based control scheme for redundant robots, where the redundant configuration is explored to achieve a safer and more efficient human-robot collaboration. In order to guarantee safety, a damping model is formulated in the null space of the robot end-effector to regulate the dynamic relationship between human effort and the robot's joint velocity. Such a formulation allows the human to get involved at any time to deal with unforeseen changes without affecting the main task of the robot end-effector, which then guarantees efficiency. A series of online adaption laws are also designed to deal with unknown parameters in the vision-based control system, such that the convergence of task error to zero is achieved in the presence of uncalibrated cameras. The stability of the closed-loop system, in both task space and null space, is rigorously proved. Experimental results on an AR-guided robot manipulator are presented to validate the performance of the proposed controller in different scenarios. 

\section{Preliminaries}
\subsection{Null Space of Redundant Robot}
The forward kinematic relationship for a robot manipulator can be described as
\begin{eqnarray}
&\bm r=\bm h(\bm q),
\end{eqnarray}
where $\bm r\hspace{-0.05cm}\in\hspace{-0.05cm}\Re^m$ denotes the feature's position in Cartesian space (e.g., robot end-effector), 
$\bm q\hspace{-0.05cm}\in\hspace{-0.05cm}\Re^n$ is the vector of joint angles, and $\bm h(\cdot)\hspace{-0.1cm}\in\hspace{-0.1cm}\Re^n\rightarrow \Re^m$ denotes the function of forward kinematics. 

Next, the relationship between the velocity of the feature in Cartesian space and the joint-space velocity is\cite{sensoryfeedbackbook}
\begin{eqnarray}
&\dot{\bm r}=\bm J(\bm q)\dot{\bm q},\label{kinematic1_}
\end{eqnarray}
where $\bm J(\cdot)\hspace{-0.05cm}\in\hspace{-0.05cm}\Re^{m\times n}$ is the Jacobian matrix from joint space to Cartesian space. 

Consider a redundant robot where $n\hspace{-0.05cm}>\hspace{-0.05cm}m$. The pseudo-inverse of the Jacobian matrix is given as
\begin{eqnarray}
&\bm J^+(\bm q)\stackrel{\triangle}{=}\bm J^T(\bm q)(\bm J(\bm q)\bm J^T(\bm q))^{-1}\in\Re^{n\times m},
\end{eqnarray}
such that $\bm J(\bm q)\bm J^+(\bm q)\hspace{-0.1cm}=\hspace{-0.1cm}\bm I_m$, where $\bm I_m\hspace{-0.05cm}\in\hspace{-0.05cm}\Re^{m\times m}$ is an identity matrix. This holds as long as $\bm J(\bm q)$ is nonsingular. Now, the null-space matrix can be introduced as \cite{tro14_sadeghian}
\begin{eqnarray}
&\bm N(\bm q)\stackrel{\triangle}{=}\bm I_n-\bm J^+(\bm q)\bm J(\bm q)\in\Re^{n\times n},
\end{eqnarray}
where $\bm I_n\hspace{-0.1cm}\in\hspace{-0.1cm}\Re^{n\times n}$ also represents an identity matrix.
Note that
$\bm J(\bm q)\bm N(\bm q)\hspace{-0.05cm}=\hspace{-0.05cm}\bm 0$,  $\bm N(\bm q)\bm J^+(\bm q)\hspace{-0.05cm}=\hspace{-0.05cm}\bm 0$, and $\bm N^2(\bm q)\hspace{-0.05cm}=\hspace{-0.05cm}\bm N(\bm q)$. 

\subsection{Vision Space and Camera Model}
This paper considers a vision-based robot manipulator, where the manipulation task is specified in vision space. Specifically, the feature's velocity in vision space can be related to the velocity of robot end-effector in Cartesian space, under an eye-to-hand configuration as
\begin{eqnarray}
&\dot{\bm x}=\frac{1}{z(\bm q)}\bm J_s(\bm r)\dot{\bm r},\label{kinematic1}
\end{eqnarray}
where $\bm x\hspace{-0.05cm}\in\hspace{-0.05cm}\Re^2$ is the feature's position in vision space (i.e., the pixel coordinate), $\bm J_s(\bm r)\hspace{-0.05cm}\in\hspace{-0.05cm}\Re^{2\times m}$ is the image Jacobian matrix, and $z(\bm q)$ is a scalar, representing the depth of the feature with respect to the focal plane, depending on the joint configuration.

Note that (\ref{kinematic1}) can be represented as
\begin{eqnarray}
&z(\bm q)\dot{\bm x}=\bm J_s(\bm r)\dot{\bm r},\label{kinematic2}
\end{eqnarray}
where both sides can be parameterized respectively as \cite{sensoryfeedbackbook}
\begin{eqnarray}
&z(\bm q)\dot{\bm x}=\bm Y_z(\dot{\bm x}, \bm q)\bm\theta_z,\label{paraDepth}\\
&\bm J_s(\bm r)\dot{\bm r}=\bm Y_k(\dot{\bm r}, \bm r)\bm\theta_k.\label{kinematic3}
\end{eqnarray}
The $\bm Y_z(\cdot)\hspace{-0.05cm}\in\hspace{-0.05cm}\Re^{2\times n_z}$, $\bm Y_k(\cdot)\hspace{-0.05cm}\in\hspace{-0.05cm}\Re^{2\times n_k}$ are regressor matrices. $\bm \theta_z\hspace{-0.05cm}\in\hspace{-0.05cm}\Re^{n_z}$ is the vector of parameters for the depth. $\bm \theta_k\hspace{-0.05cm}\in\hspace{-0.05cm}\Re^{n_k}$ is the vector of intrinsic and extrinsic parameters of the camera. $n_z$ and $n_k$ denote the dimension. Note that the actual values of $\bm\theta_k$ and $\bm\theta_z$ are unknown if the camera is uncalibrated.
\\\\
\hspace{-0.3cm}\textbf{Problem Statement}:
{\em Consider the kinematic control problem of a redundant robot, where the control input is executed on the joint velocity as
\begin{eqnarray}
&\dot{\bm q}=\bm u,\label{kinematicModel}
\end{eqnarray}
where $\bm u\in\Re^n$ denotes the input.
The objective is to design the input to achieve both the main task in vision space and the interaction task in null space.}

\section{Vision-Based Control with Null-Space interaction}
\begin{center}
\begin{figure}[!h]
\centering
\includegraphics[width=2.5in]{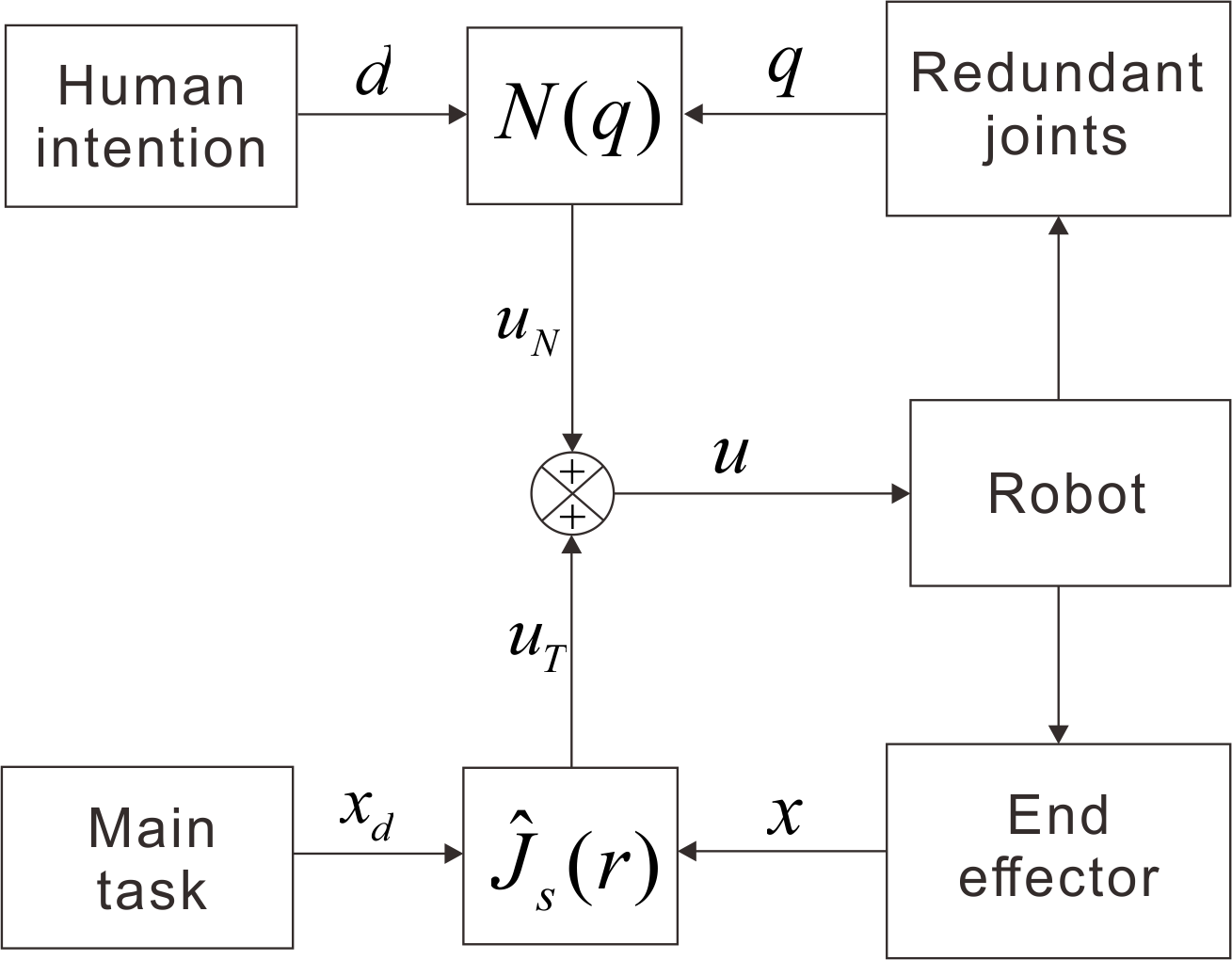}
\caption{Block diagram of the proposed control scheme. In the presence of an uncalibrated camera, the exact information about the image Jacobian matrix is unknown and denoted as $\hat{\bm J}_s(\bm r)$. In vision space, the robot end-effector is controlled to move to the desired position, that is, $\bm x\rightarrow\bm x_d$; In null space, the redundant joints are controlled to suit human intention under the desired damping model, that is, $c_d\dot{\bm q}=\bm d$, where $c_d$ is a positive constant and $\bm d$ denotes the control efforts from a human.}\label{bd}
\end{figure}
\end{center}

In this section, an adaptive vision-based controller with null-space interaction is proposed, and its structure is illustrated in Fig. \ref{bd}. In vision space, the robot end-effector is controlled to move to the desired position in the presence of an uncalibrated camera. In null space, the redundant joints are controlled to suit the human intention under the desired damping model, such that humans can get involved in dealing with unforeseen changes during the manipulation.
\vspace{-1.0cm}

\subsection{Controller Development}
First, the control objective in sensory space is to ensure the convergence of $\bm x\hspace{-0.05cm}\rightarrow\hspace{-0.05cm}\bm x_d$ where $\bm x_d\hspace{-0.05cm}\in\hspace{-0.05cm}\Re^2$ is the desired position in vision space, specifying the main task.

When the robot is working in an uncalibrated environment, the exact information about the camera model is usually unknown (e.g., the unknown depth information). While calibration techniques may be implemented to obtain the knowledge of the model, identifying or solving parameters requires effort. Even after considerable works on the modeling, some uncertainty may still exist, affecting the stability of the control system. Moreover, calibration must be repeated when the robotic system is migrated to similar scenarios. The aforementioned have motivated us to consider the uncalibrated camera as a general situation in this paper.   

In the presence of the uncalibrated camera, the unknown depth and image Jacobian matrix are represented as $\hat z(\bm q)$ and $\hat{\bm J}_s(\bm r)$, respectively. From (\ref{paraDepth}) and (\ref{kinematic3}) it is obtained that
\begin{eqnarray}
&\hat z(\bm q)\dot{\bm x}=\bm Y_z(\dot{\bm x}, \bm q)\hat{\bm\theta}_z,\label{depthprop}\\
&\hat{\bm J}_s(\bm r)\dot{\bm r}=\bm Y_k(\dot{\bm r}, \bm r)\hat{\bm\theta}_k,\label{kinematicUnknown}
\end{eqnarray}
where $\hat{\bm\theta}_z$ and $\hat{\bm\theta}_k$ denote the estimates of ${\bm\theta}_z$ and ${\bm\theta}_k$, respectively.

Now, the vision-space control term is proposed as
\begin{eqnarray}
&\bm u_T\hspace{-0.05cm}=\hspace{-0.05cm}-\hat{z}(\bm q)\bm J^+(\bm q)\hat{\bm J}_s^+(\bm r)\bm K_p(\bm x\hspace{-0.05cm}-\hspace{-0.05cm}\bm x_d),\label{xp}
\end{eqnarray}
where $\bm K_p\hspace{-0.1cm}\in\hspace{-0.1cm}\Re^{2\times 2}$ represents the control gain which is a diagonal and positive-definite matrix, $\hat{\bm J}_s^+(\bm r)$ is the pseudo-inverse of $\hat{\bm J}_s(\bm r)$. In the vision-space control term (\ref{xp}), the uncertain parameters of depth and image Jacobian matrix in (\ref{xp}) are updated by the following adaptation laws
\begin{eqnarray}
&\dot{\hat{\bm\theta}}_z=-\hat{z}^{-1}(\bm q)\bm L_z \bm Y_z^T(\dot{\bm x}, \bm q)(\bm x-\bm x_d),\label{updateThetaz}\\
&\dot{\hat{\bm\theta}}_k=\hat{z}^{-1}(\bm q)\bm L_k \bm Y_k^T(\dot{\bm r}, \bm r)(\bm x-\bm x_d),\label{updateThetak}
\end{eqnarray}
where $\bm L_z\hspace{-0.1cm}\in\hspace{-0.1cm}\Re^{n_z\times n_z}, \bm L_k\hspace{-0.1cm}\in\hspace{-0.1cm}\Re^{n_k\times n_k}$ are diagonal and positive-definite, which govern the convergence of the estimated parameters. The updating laws in (\ref{updateThetaz}) and (\ref{updateThetak}) are driven by the vision-space error $\bm x\hspace{-0.05cm}-\hspace{-0.05cm}\bm x_d$, which is always activated until the end-effector reaches the desired position, i.e., $\bm x\hspace{-0.05cm}\rightarrow\hspace{-0.05cm}\bm x_d$.

Next, the objective of null-space control is to achieve the desired damping model as
\begin{eqnarray}
{\bm N}(\bm q)(c_d\dot{\bm q}-\bm d)=\bm 0,\label{desiredIm}
\end{eqnarray}
where $c_d$ is a positive constant which denotes the desired damping factor, which is positive, and $\bm d$ specifies the human intention. The model (\ref{desiredIm}) damps the redundant joints, and hence makes them compliant to the input $\bm d$. Note that the damping parameter can be adjusted according to the specific task.

\begin{center}
\begin{figure}[!h]
\centering
\includegraphics[width=2.7in]{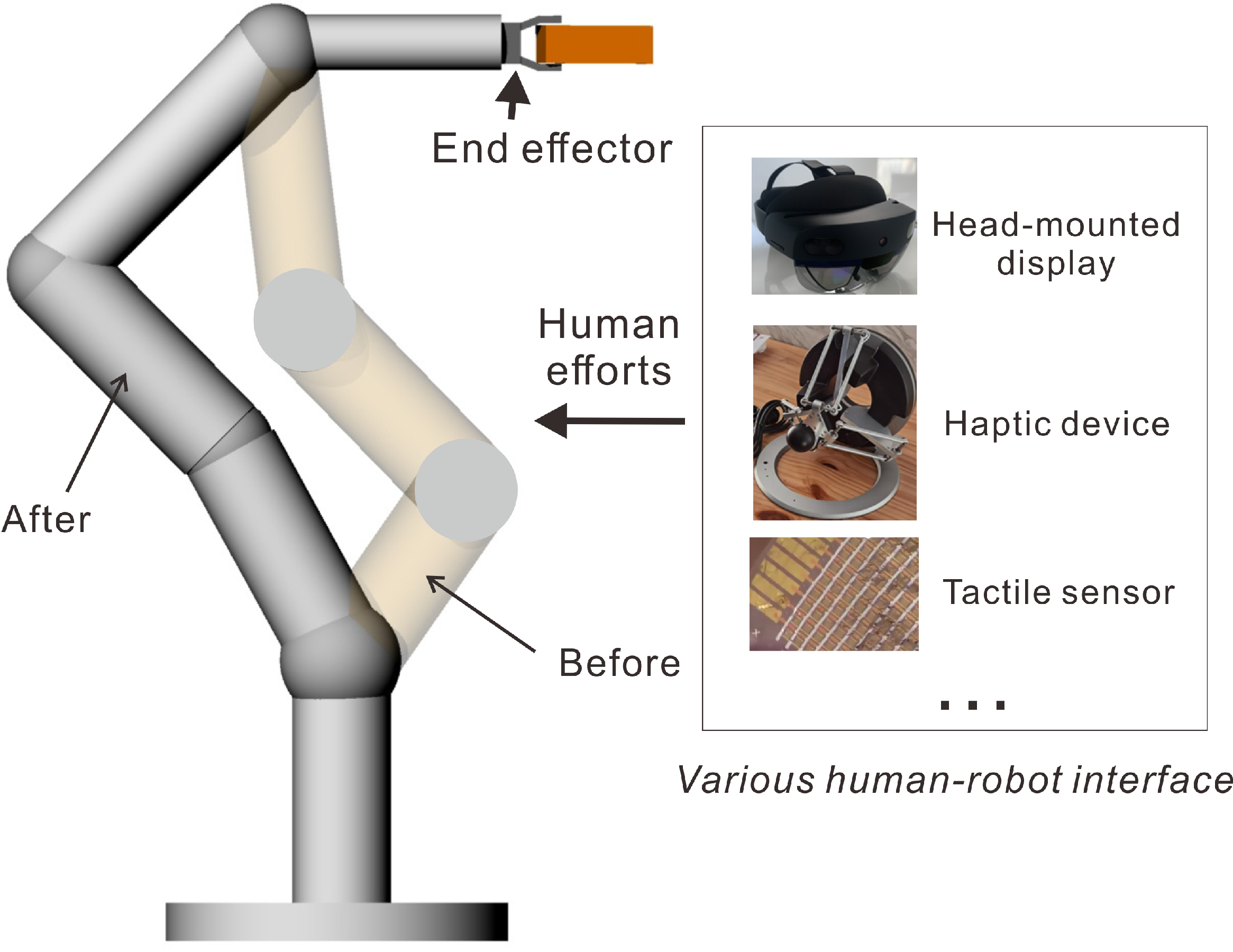}
\caption{The human can input via various interfaces, e.g., head-mounted display, haptic device, or tactile sensor, to shape the redundant configuration without affecting the robot end-effector.}\label{ex}
\end{figure}
\end{center}

Now, the null-space interactive control term can be proposed as
\begin{eqnarray}
&\bm u_N={\bm N}(\bm q)(c^{-1}_d\bm d).\label{visualFeedback}
\end{eqnarray}
Note that the vector $\bm d$ defines a general descriptor for human control efforts, which can be specified in different forms (see Fig. \ref{ex}). For example, it represents the interaction force or torque when the human directly pushes or pulls the robot; it can be the command sent from a joystick or AR device, where the human is involved in a teleoperated way. \\
\textbf{Remark 1}: The key novelty of the proposed null-space interactive control is to make the redundant joints passive and hence allow the human to exert additional control efforts for collaboration. 
As shown in Fig. \ref{ex}, the control efforts from humans shapes the redundant joints without affecting the main task of the end-effector. This is useful when the overall body may need to be controlled to avoid dynamic obstacles or suit the environment. 
\vspace{-0.4cm}
\vspace{-0.5cm}

\subsection{Stability Analysis}
The overall control input is the summation of both control terms, that is,
\begin{eqnarray}
&\bm u=\bm u_T +\bm u_N. \label{CaFun}
\end{eqnarray}
Then, substituting (\ref{CaFun}) into (\ref{kinematicModel}) yields
\begin{eqnarray}
&\dot{\bm q}\hspace{-0.05cm}=\hspace{-0.05cm}-\hat{z}(\bm q)\bm J^+(\bm q)\hat{\bm J}_s^+(\bm r)\bm K_p(\bm x-\bm x_d)\nonumber\\
&+{\bm N}(\bm q)(c_d^{-1}\bm d).\label{closed}
\end{eqnarray}
Multiplying both sides of (\ref{closed}) with $\bm J(\bm q)$, the closed-loop equation in Cartesian space is obtained as
\begin{eqnarray}
\dot{\bm r}=-\hat{z}(\bm q)\hat{\bm J}_s^+(\bm r)\bm K_p(\bm x\hspace{-0.05cm}-\hspace{-0.05cm}\bm x_d).\label{temp2}
\end{eqnarray}
Multiplying both sides of (\ref{closed}) with $\bm N(\bm q)$ and using the property of $\bm N^2(\bm q)\hspace{-0.05cm}=\hspace{-0.05cm}\bm N(\bm q)$, the closed-loop equation in null space is obtained as
\begin{eqnarray}
\bm N(\bm q) \dot{\bm q}=\bm N(\bm q)(c_d^{-1}\bm d),\label{temp2Null}
\end{eqnarray}
which clearly shows the realization of the desired damping model (\ref{desiredIm}).

Next, multiplying both sizes of (\ref{temp2}) with $\hat{z}^{-1}(\bm q)\hat{\bm J}_s(\bm r)$, it is obtained that
\begin{eqnarray}
&\hat{z}^{-1}(\bm q)\hat{\bm J}_s(\bm r)\dot{\bm r}=-\bm K_p(\bm x\hspace{-0.05cm}-\hspace{-0.05cm}\bm x_d).\label{Newtemp2}
\end{eqnarray}

Note that
\begin{eqnarray}
&\hat{z}^{-1}(\bm q)\hat{\bm J}_s(\bm r)\dot{\bm r}=\hat{z}^{-1}(\bm q)\hat{\bm J}_s(\bm r)\dot{\bm r}-\hat z^{-1}(\bm q)z(\bm q)\dot{\bm x}\nonumber\\
&+\hat z^{-1}(\bm q)z(\bm q)\dot{\bm x}-\dot{\bm x}+\dot{\bm x}\nonumber\\
&=\hat{z}^{-1}(\bm q)[\hat{\bm J}_s(\bm r)\dot{\bm r}-z(\bm q)\dot{\bm x}]\nonumber\\
&+\hat z^{-1}(\bm q)[z(\bm q)\dot{\bm x}-\hat z(\bm q)\dot{\bm x}]+\dot{\bm x},\label{temp3}
\end{eqnarray}

By referring to (\ref{kinematic2}), (\ref{paraDepth}), and (\ref{kinematic3}), the above equation can be written as
\begin{eqnarray}
&\hat{z}^{-1}(\bm q)[\hat{\bm J}_s(\bm r)\dot{\bm r}-z(\bm q)\dot{\bm x}]\nonumber\\
&+\hat z^{-1}(\bm q)[z(\bm q)\dot{\bm x}-\hat z(\bm q)\dot{\bm x}]+\dot{\bm x}\nonumber\\
&=-\hat{z}^{-1}(\bm q)[{\bm J}_s(\bm r)\dot{\bm r}-\hat{\bm J}_s(\bm r)\dot{\bm r}]\nonumber\\
&+\hat z^{-1}(\bm q)[z(\bm q)\dot{\bm x}-\hat z(\bm q)\dot{\bm x}]+\dot{\bm x}\nonumber\\
&=\dot{\bm x}-\hat{z}^{-1}(\bm q)[\bm Y_k(\dot{\bm r}, \bm r)\Delta\bm\theta_k-\bm Y_z(\dot{\bm x}, \bm q)\Delta\bm\theta_z],\label{temp41}
\end{eqnarray}
where $\Delta\bm\theta_k=\bm\theta_k-\hat{\bm\theta}_k$ and $\Delta\bm\theta_z=\bm\theta_z-\hat{\bm\theta}_z$.

Substituting (\ref{temp41}) into (\ref{temp2}), we have
\begin{eqnarray}
&\dot{\bm x}\hspace{-0.05cm}=\hspace{-0.05cm}\hat{z}^{-1}(\bm q)[\bm Y_k(\dot{\bm r}, \bm r)\Delta\bm\theta_k-\bm Y_z(\dot{\bm x}, \bm q)\Delta\bm\theta_z]\nonumber\\
&-\bm K_p(\bm x\hspace{-0.05cm}-\hspace{-0.05cm}\bm x_d).\label{atemp4}
\end{eqnarray}

We are now in the position to state the following theorem.\\\\
\textbf{Theorem}:
{\em The proposed adaptive control scheme described by (\ref{xp}), (\ref{updateThetaz}), (\ref{updateThetak}), (\ref{visualFeedback}), and (\ref{CaFun})
	ensures the convergence of the position error in vision space to zero and also the realization of the desired damping model in null space, in the presence of the uncalibrated camera.
}

\begin{proof}
To prove the stability, a Lyapunov candidate is proposed as
\begin{eqnarray}
&V=\frac{1}{2}(\bm x-\bm x_d)^T(\bm x-\bm x_d)+\frac{1}{2}\Delta\bm\theta_k^T\bm L_k^{-1}\Delta\bm\theta_k\nonumber\\
&+\frac{1}{2}\Delta\bm\theta_z^T\bm L_z^{-1}\Delta\bm\theta_z.\label{lyafun}
\end{eqnarray}

Differentiating (\ref{lyafun}) with respect to time yields
\begin{eqnarray}
&\dot V=(\bm x-\bm x_d)^T\dot{\bm x}-\Delta\bm\theta_k^T\bm L_k^{-1}\dot{\hat{\bm\theta}}_k-\Delta\bm\theta_z^T\bm L_z^{-1}\dot{\hat{\bm\theta}}_z,\label{dlyafun}
\end{eqnarray}
where $\dot{\bm x}_d\hspace{-0.05cm}\equiv\hspace{-0.05cm}\bm 0$. Then, substituting (\ref{atemp4}) into (\ref{dlyafun}) yields
\begin{eqnarray}
&\dot V=(\bm x-\bm x_d)^T\{-\bm K_p(\bm x\hspace{-0.05cm}-\hspace{-0.05cm}\bm x_d)\\
&+\hat{z}^{-1}(\bm q)[\bm Y_k(\dot{\bm r}, \bm r)\Delta\bm\theta_k-\bm Y_z(\dot{\bm x}, \bm q)\Delta\bm\theta_z]\}\nonumber\\
&-\Delta\bm\theta_k^T\bm L_k^{-1}\dot{\hat{\bm\theta}}_k-\Delta\bm\theta_z^T\bm L_z^{-1}\dot{\hat{\bm\theta}}_z.
\end{eqnarray}

Substituting the update laws (\ref{updateThetaz}) and (\ref{updateThetak}) into the above equation yields
\begin{eqnarray}
&\dot V=-(\bm x-\bm x_d)^T\bm K_p(\bm x-\bm x_d)\leq 0. \label{temp4}
\end{eqnarray}

Since $V\hspace{-0.05cm}>\hspace{-0.05cm}0$ and $\dot V\hspace{-0.05cm}\leq\hspace{-0.05cm}0$, $V$ is bounded. The boundedness of $V$ ensures the boundedness of $\bm x-\bm x_d$, $\Delta\bm\theta_z$, and $\Delta\bm\theta_k$. Then, from (\ref{temp2}), it is derived that $\dot{\bm r}$ is bounded, indicating that $\dot{\bm x}$ is also bounded from (\ref{kinematic1}). Hence, $\bm x-\bm x_d$ is uniformly continuous. It can be derived from (\ref{temp4}) that $(\bm x-\bm x_d)\hspace{-0.05cm}\in\hspace{-0.05cm}L_2(0, \infty)$ (i.e., $L_2$ boundedness). Then, according to \cite{aribook,slotine}, $\bm x-
\bm x_d\rightarrow\bm 0$, that is, $\bm x\rightarrow
\bm x_d$ as $t\rightarrow\infty$. As analyzed previously, (\ref{temp2Null}) has guaranteed the realization of the desired damping model. Therefore, both the tasks in the sensory space and null space are achieved.
\end{proof}

\begin{center}
\begin{figure}[!tb]
\centering
  \subfigure[]{ 
    \includegraphics[width=7cm]{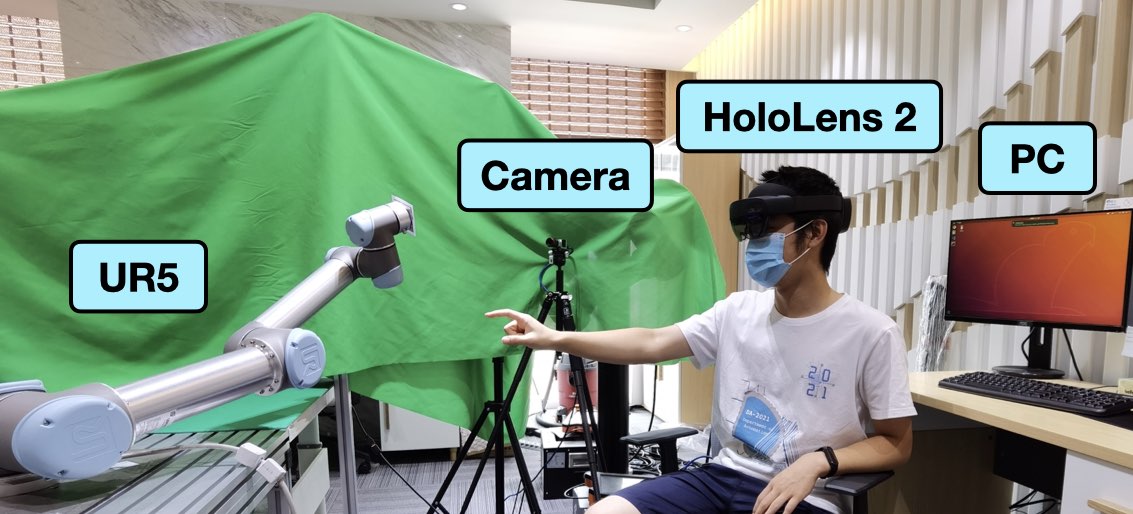}
  }
  \subfigure[]{ 
    \includegraphics[width=7cm]{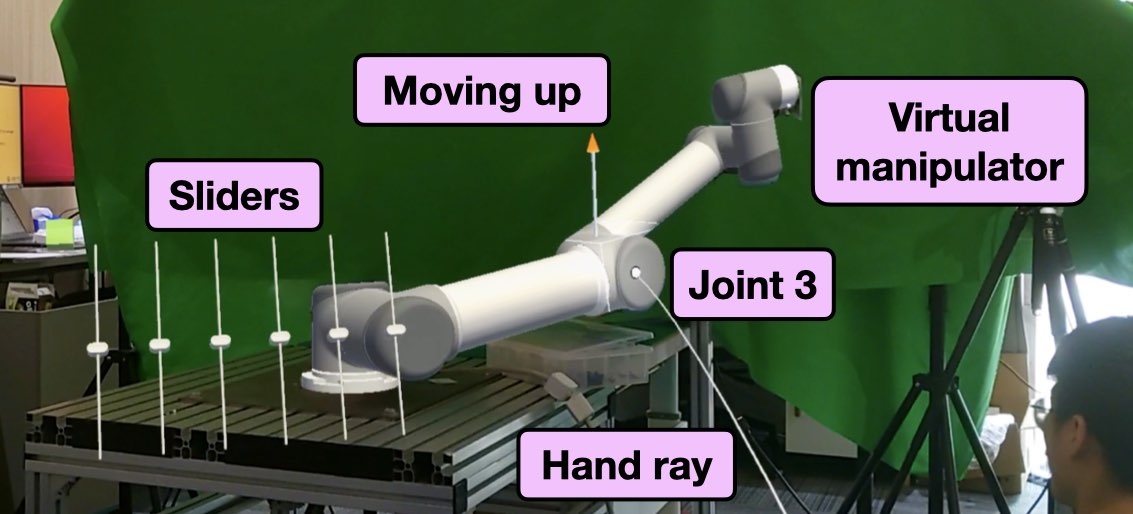} \label{exp-setup}
  }
\vspace{-0.3cm}
\caption{(a) Experimental setup of an AR-guided robot manipulator (b) A snapshot of the human-robot interface deployed on HoloLens 2. The sliders were placed on the left side, and the virtual manipulator model was on the right side, overlapped with the real manipulator. The human operator was pointing at joint 3 and giving a sign of moving up. The virtual manipulator model could also be placed somewhere else according to the situations.}\label{expsetup}
\vspace{-5mm}
\end{figure}
\end{center}

\begin{figure*} [!tb]
  \centering 
  \subfigure[]{ 
    \includegraphics[width=4.3cm]{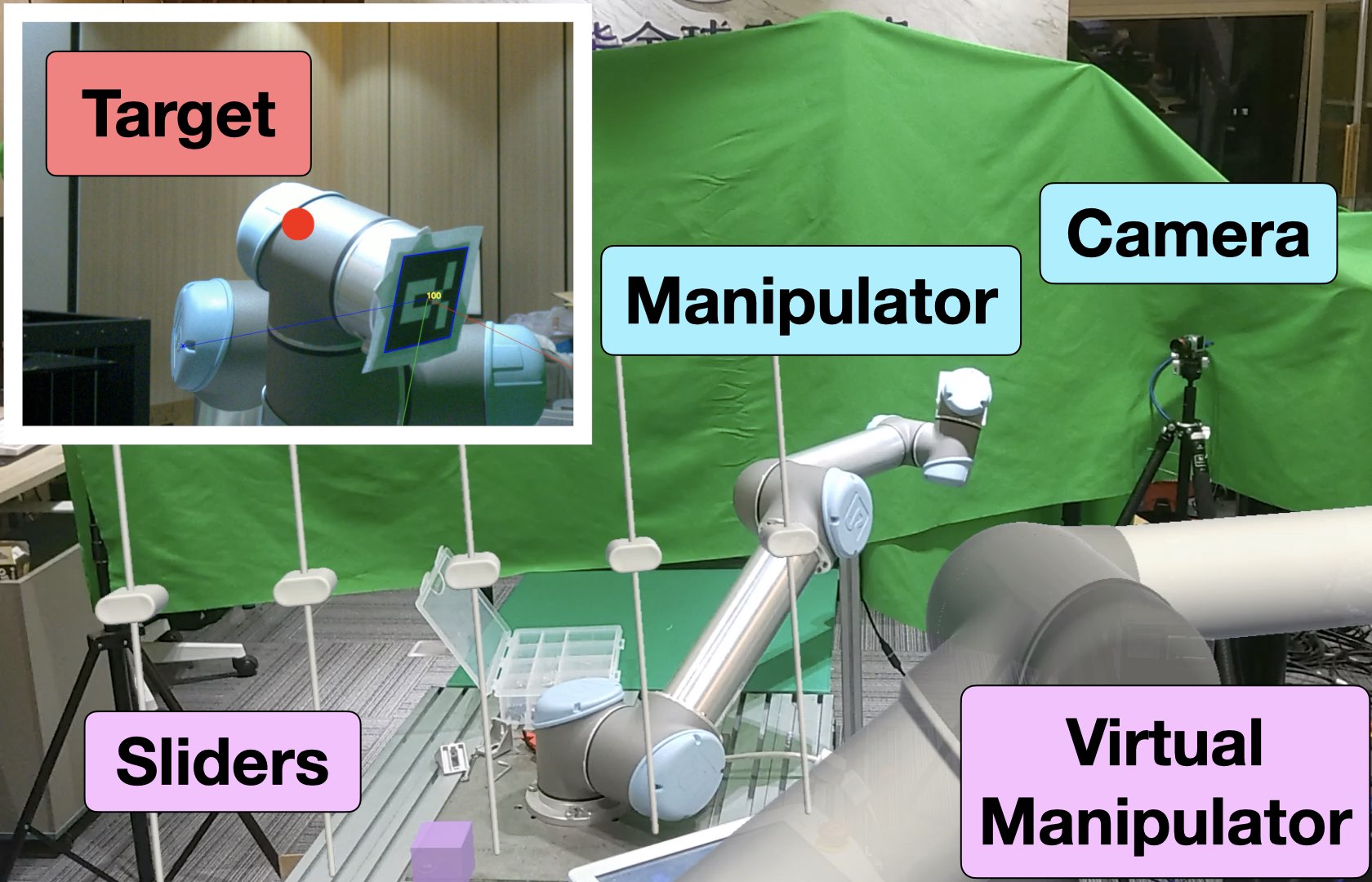}
  }
  \hspace{-0.4cm}
  \subfigure[]{ 
    \includegraphics[width=4.3cm]{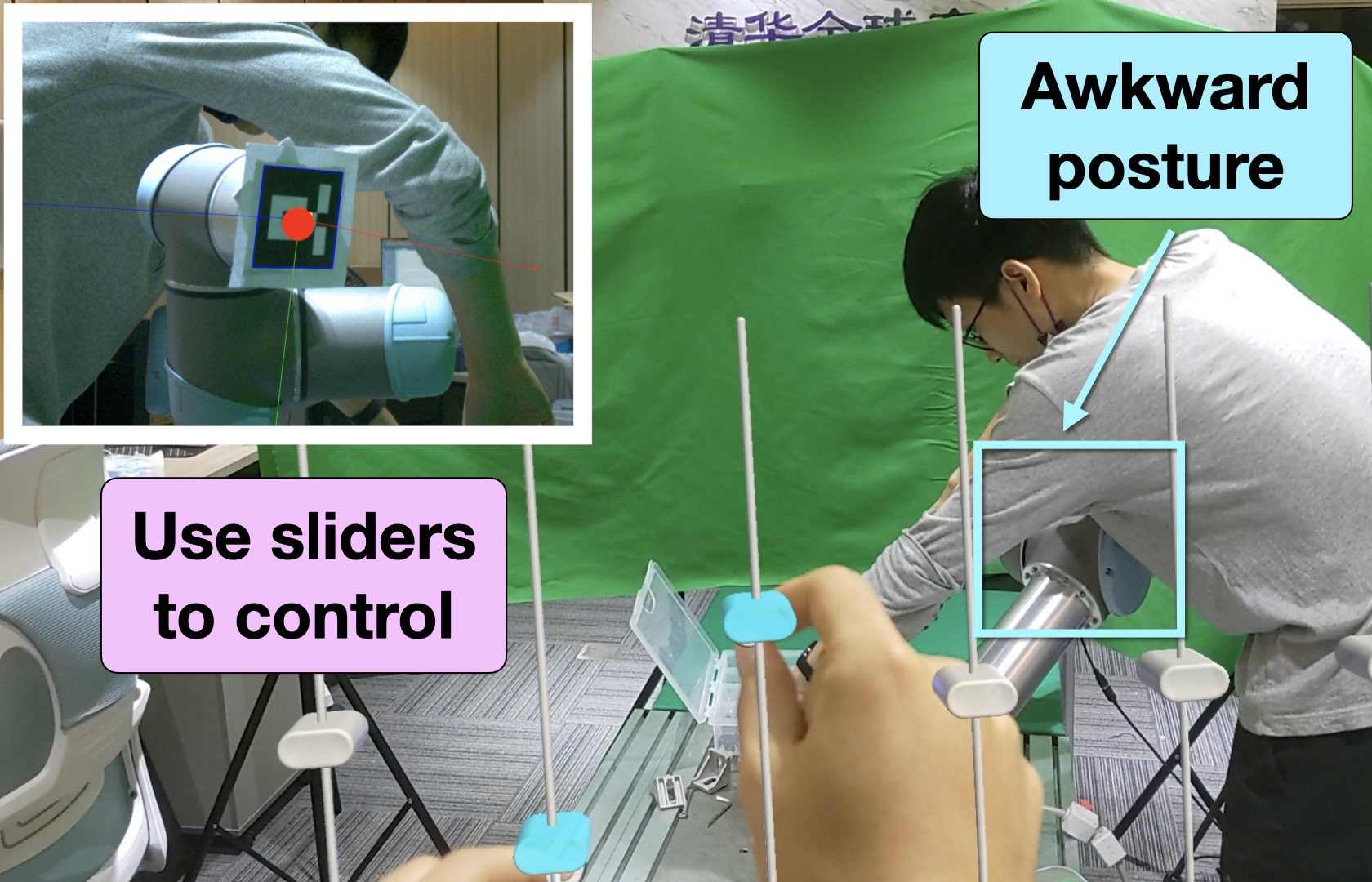}
  } 
  \hspace{-0.4cm}
  \subfigure[]{ 
    \includegraphics[width=4.3cm]{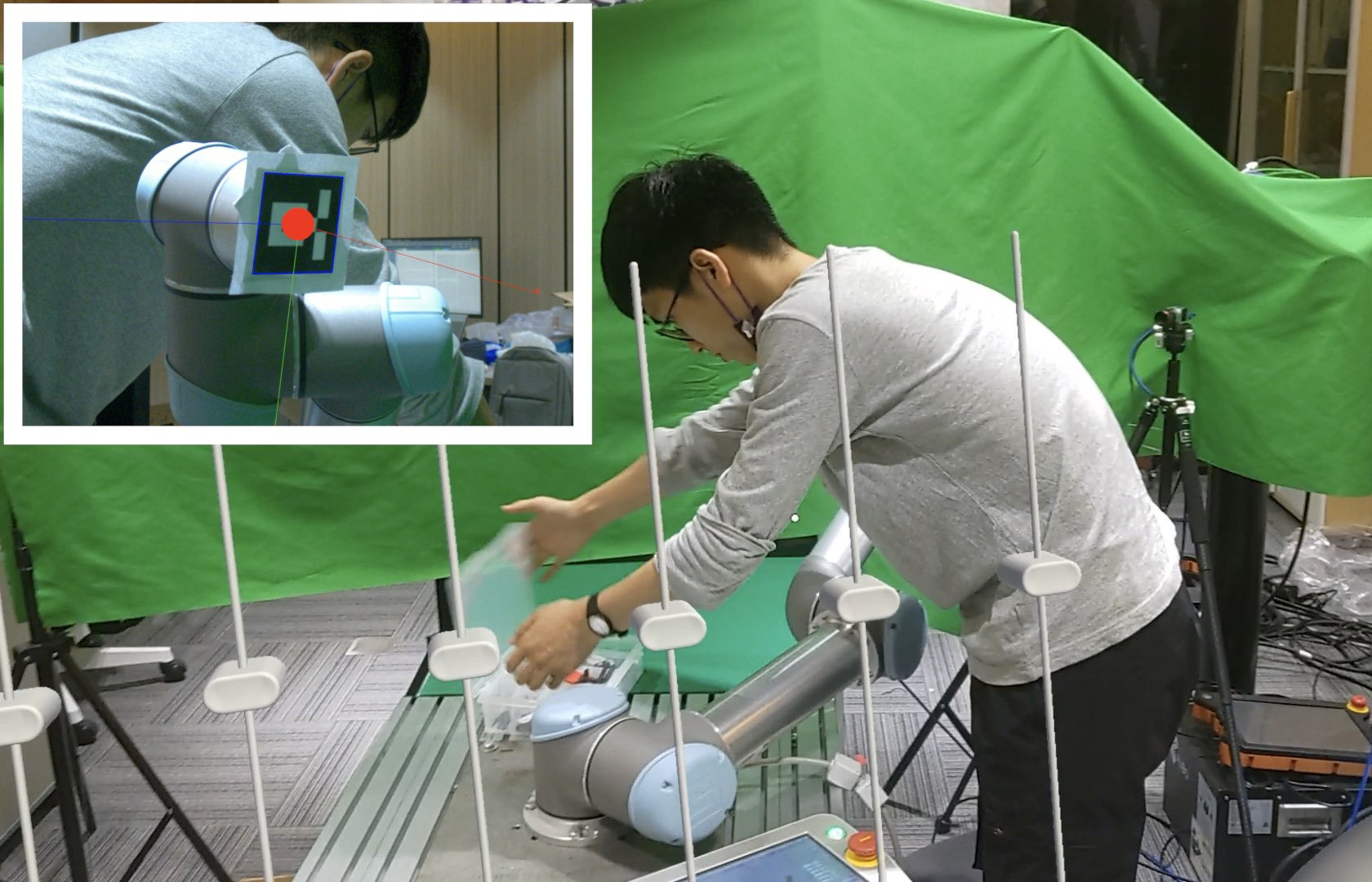} 
  }
  \hspace{-0.4cm}
  \subfigure[]{ 
    \includegraphics[width=4.3cm]{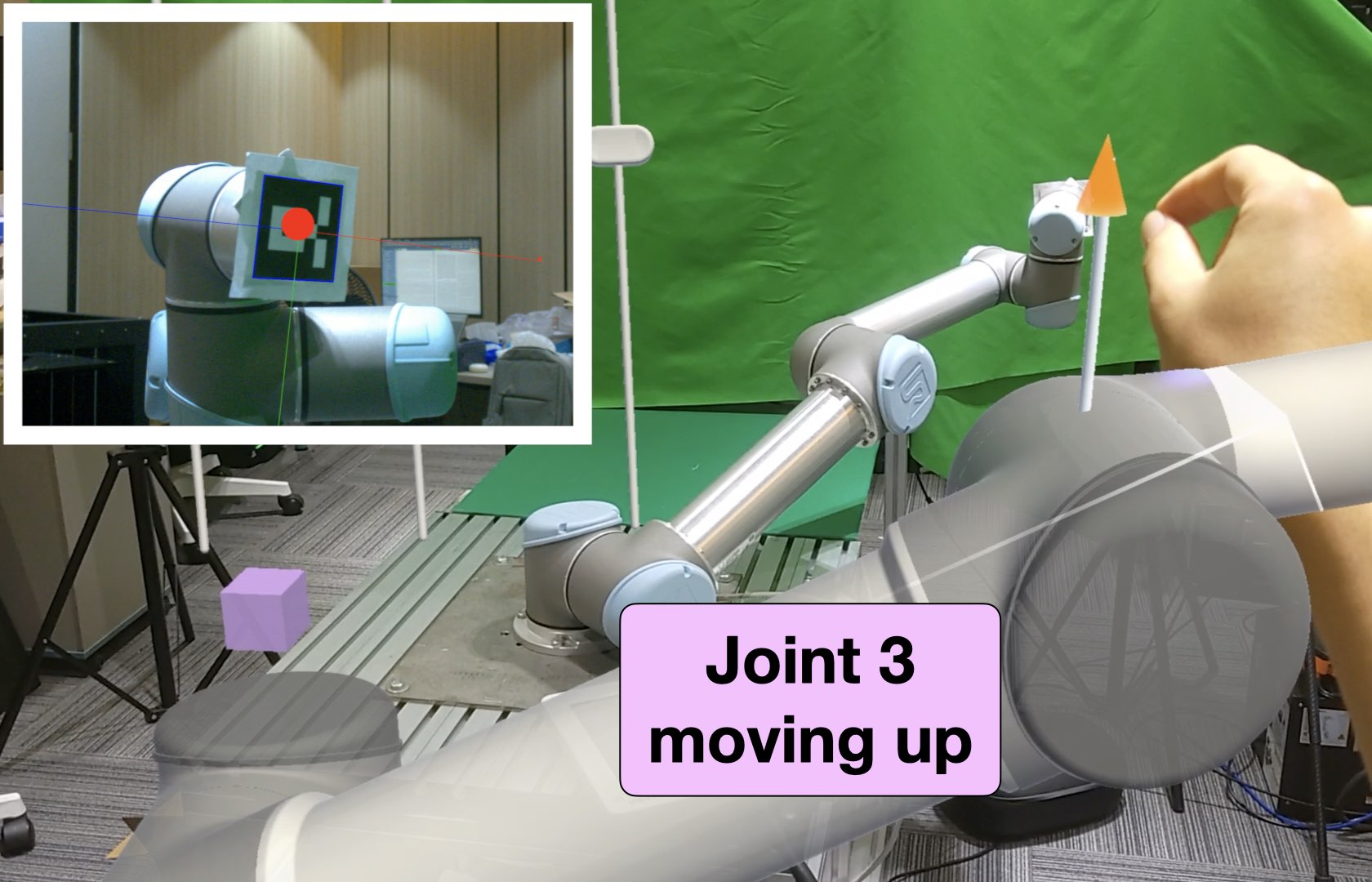} 
  } 
  \vspace{-3mm}
  \caption{Snapshots of task 1. The manipulator was controlled to move to the desired point in vision space, while the operator exerted additional control efforts to make another human subject more comfortable via the AR interface.
    (a) $t = 0.0\hspace{0.2mm} {\rm s}$: The end-effector started at the lower-right corner of the camera frame, and would move to the desired point at the center in $1.6\hspace{0.2mm} {\rm s}$.
    (b) $t = 6.0 \hspace{0.2mm} {\rm s}$: The operator began to drag sliders corresponding to joint 2 and joint 3 to change the manipulator's configuration because another human subject entered the workspace and stretched his body in an uncomfortable pose to fetch tools.
    (c) $t = 20.1 \hspace{0.2mm} {\rm s}$: The manipulator adjusted its pose according to the command, while the position of the end-effector in vision space remained the same. Meanwhile, it enabled the human subject to have a safer and more comfortable workspace.
    (d) $t = 30.3 \hspace{0.2mm} {\rm s}$: After the subject left, the robot's joint moved back to its original position according to the operator's command.}
    \label{e1s}
    \vspace{-3mm}
\end{figure*}

\begin{figure}[!tb]
\centering 
  \subfigure[]{
    \includegraphics[height=3.5cm]{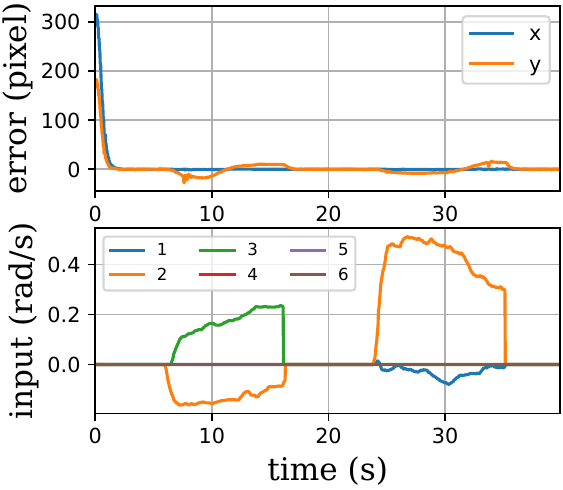}
    }
    \subfigure[]{ 
    \includegraphics[height=3.5cm]{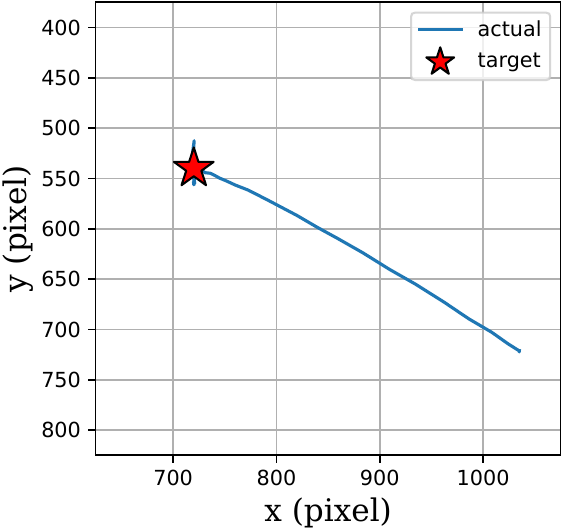} 
  } 
  \caption{Results of task 1:
    (a) top: The position errors of the end-effector in vision space; bottom: The human's control efforts applied to the joints.
    (b) The trajectory of the end-effector in vision space. }
        \label{e1_te}
\end{figure}

\begin{figure}[!tb]
\centering 
  \subfigure[]{
    \includegraphics[height=3.5cm]{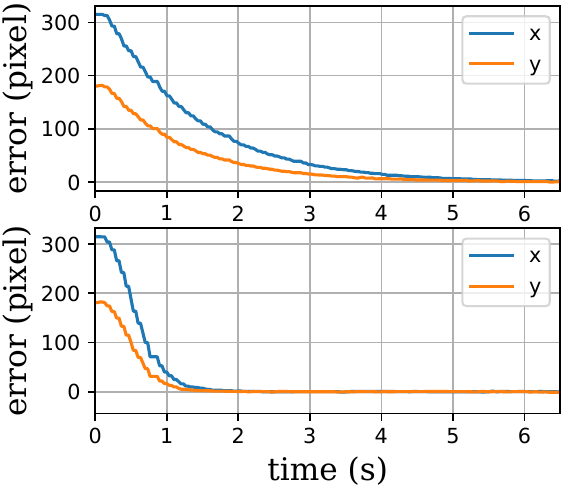}
    }
    \subfigure[]{ 
    \includegraphics[height=3.5cm]{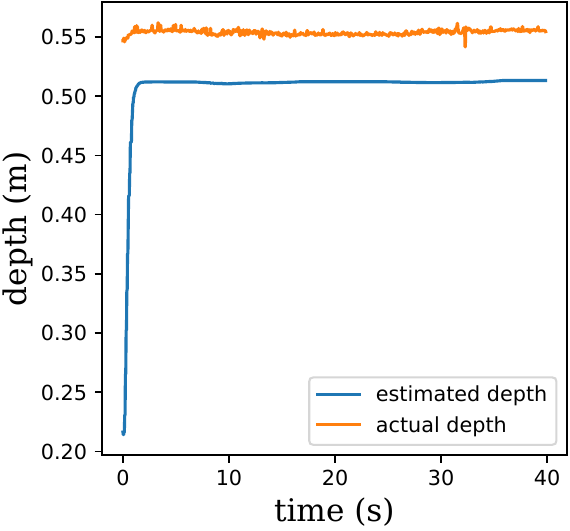} 
  } 
  \caption{Results of ablation study:
    (a) The position errors of the end-effector in vision space during the first 6.5s, without adaptation law (top) and with adaptation law (bottom).
    (b) The estimated and actual depth in the whole process of task 1. }
        \label{adaptive_comp}
        \vspace{-3mm}
\end{figure}

\vspace{-2mm}
\section{Experiment}
Experiments were carried out to validate the performance of the proposed method on an AR-guided robot manipulator, as shown in Fig. \ref{expsetup}(a). The overall system consisted of four modules: (\romannumeral1) the algorithm ran on Robot Operating System (ROS) with Ubuntu 18.04 LTS,  (\romannumeral2) the 6-degrees-of-freedom UR5 robot, whose end-effector is overlaid with an ArUco marker,  (\romannumeral3) a Basler ace acA1440-220uc camera with $1440 \times 1080$ resolution, fixed in the workspace of the robot but not calibrated, (\romannumeral4) the human-robot interface deployed on HoloLens 2, a mixed reality headset developed by Microsoft.

\subsection{Experimental Setup}
The ArUco marker was installed on the end-effector (i.e., the feature point position in the end-effector frame $^e r=[0,0,0]^T$) and set as the task feature, whose position in Cartesian space was captured by the camera and then specified in vision space. That is, the image Jacobian matrix $\hat{\bm J}_s(\bm r)$ in (\ref{kinematic1}) was a $2\times3$ matrix in our experiments. Note that the camera's depth was unknown and dependent on the specific joint configuration of the robot. It was assumed that the camera was not calibrated, and its model was initialized with some random values. The actual value can be obtained with standard calibration techniques, which was treated as the ground truth for evaluation.

The human-robot interface contained two key components (see Fig. \ref{exp-setup}), a series of sliders and a full-scale virtual manipulator. The sliders were used to define the value of the human intention $\bm d$, and the virtual manipulator was used to visualize the current configuration of the real manipulator. Both of them could be placed anywhere inside the frame, according to the operator's preference. 

When the virtual manipulator was overlapped with the real manipulator, the operator could use hand rays and air tap gestures to manipulate a certain joint; otherwise, the operator could move the virtual manipulator closer and then "grab" a certain joint. Under such framework, the human can input the control efforts in two ways, that is, 
\begin{enumerate}
\item[-] stacking the value of sliders, which was directly converted to the vector $\bm d$ in (\ref{desiredIm});
\item[-] moving a certain joint of the virtual manipulator in Cartesian space, which was represented as a vector. The vector was proportional to the command velocity. The command velocity was then projected back the joint space with the pseudo-inverse of the Jacobian matrix from base to that exact joint and represented as $\bm d$ by the end.
\end{enumerate}

\subsection{Experimental Results}
In the first experiment, the main task was to move the end-effector to the desired position in vision space. During the task, a human subject entered the robot's workspace and intended to fetch tools by stretching his body over the robot in an awkward pose.
Seeing this inconvenience, the operator dragged the sliders corresponding to joint 2 and joint 3 via the interface. Hence, the manipulator changed its configuration, improving the co-existed human's experience. 
The whole process was shown in Fig. \ref{e1s}, where the positioning task of the robot end-effector was not affected or suspended throughout the interaction.
\begin{table}[!h]
\vspace{-2mm}
\centering
\caption{Control parameters}\label{exp_para_table}
\vspace{-1mm}
\begin{tabular}{@{}ccc@{}}
\toprule
          & Task 1                   & Task 2                   \\ \midrule
$\bm K_p$ & $2\times\bm I_2$         & $5\times\bm I_2$         \\
$c_d$ & $1$                & $0.5$       \\
$\bm L_z$ & $0.001\times\bm I_{4}$ & $0.001\times\bm I_{4}$ \\
$\bm L_k$ & $0.001\times\bm I_{15}$ & $0.001\times\bm I_{15}$ \\
Target & $(720, 540)~{\rm pixel}$ & \begin{tabular}[c]{@{}c@{}}$\bm x = 720+100 \cos{\frac{\pi}{15}(t-t_0)}~{\rm pixel},$\\ $\bm y = 540+100 \sin{\frac{\pi}{15}(t-t_0)}~{\rm pixel}$\end{tabular} \\ \bottomrule
\end{tabular}
\end{table}

\begin{figure*} [!tb]
  \centering 
  \subfigure[]{
    \includegraphics[width=3.4cm]{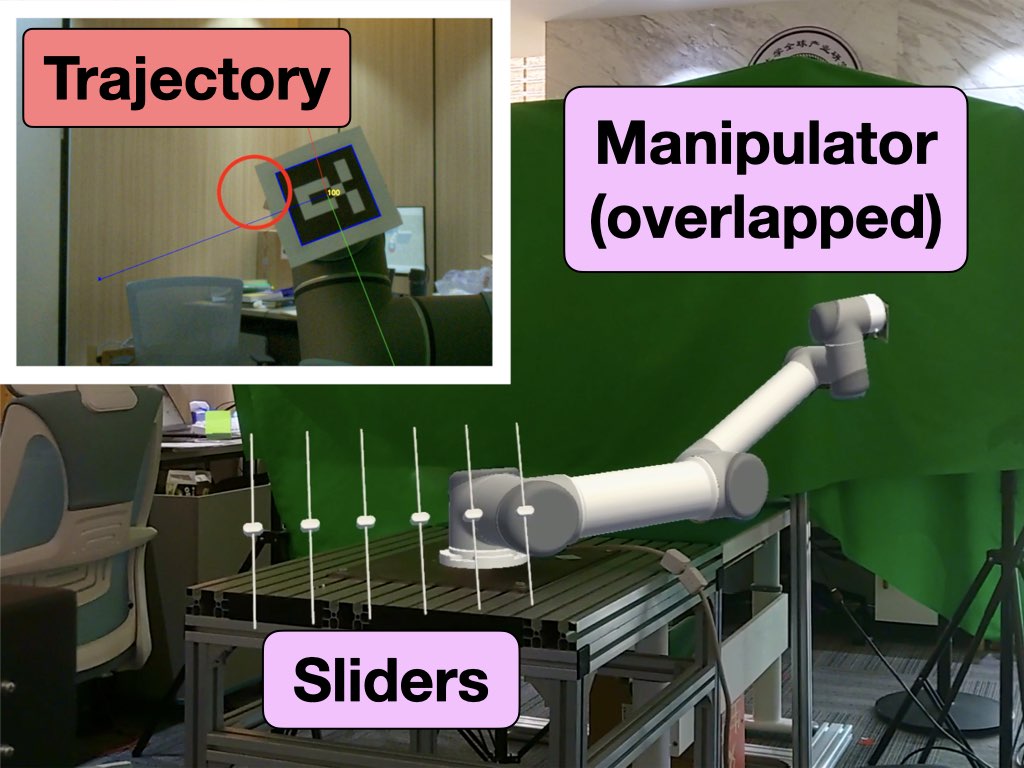} 
  }
  \hspace{-0.4cm}
  \subfigure[]{ 
    \includegraphics[width=3.4cm]{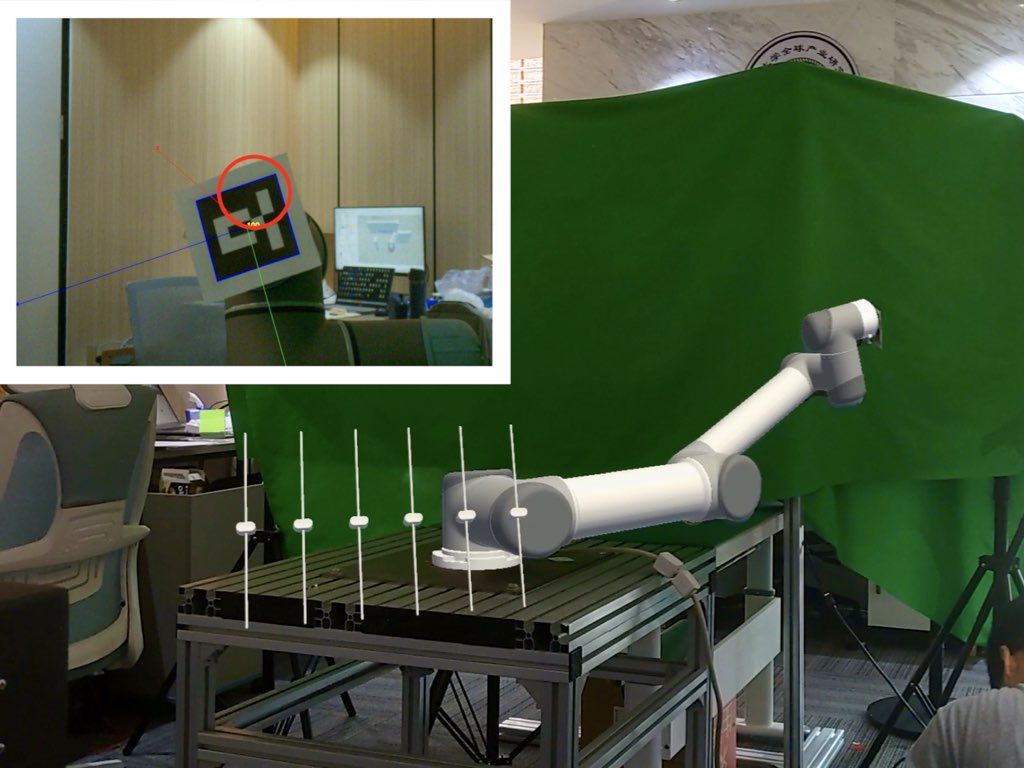} 
  } 
  \hspace{-0.4cm}
  \subfigure[]{ 
    \includegraphics[width=3.4cm]{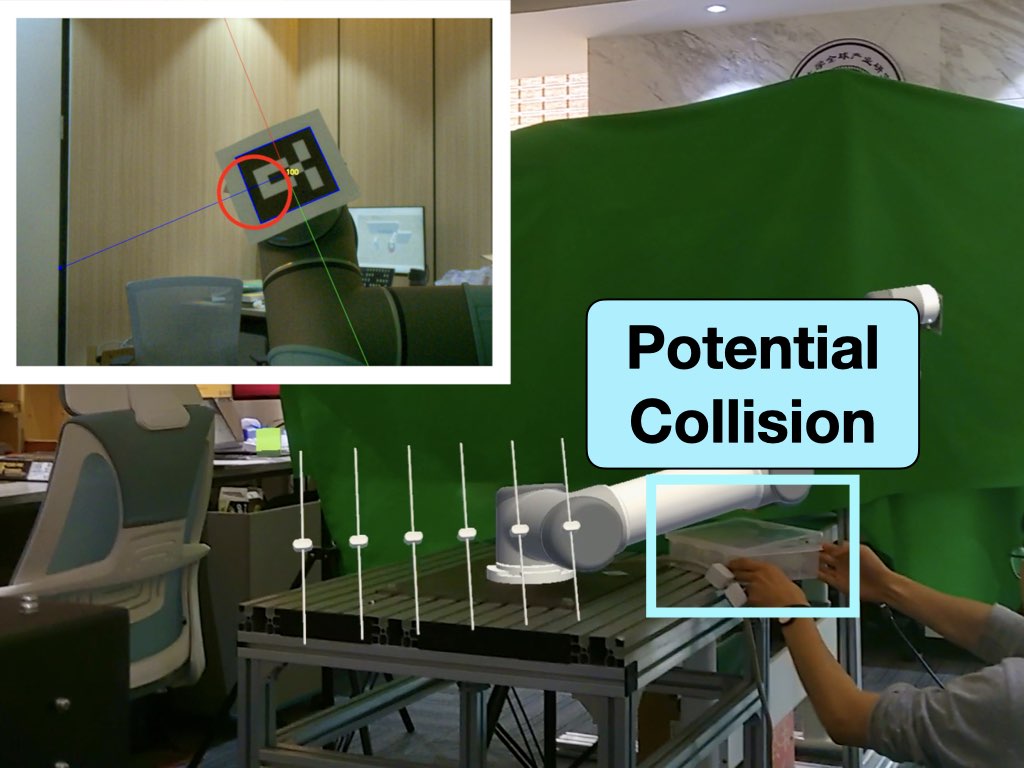} 
  }
  \hspace{-0.4cm}
  \subfigure[]{ 
    \includegraphics[width=3.4cm]{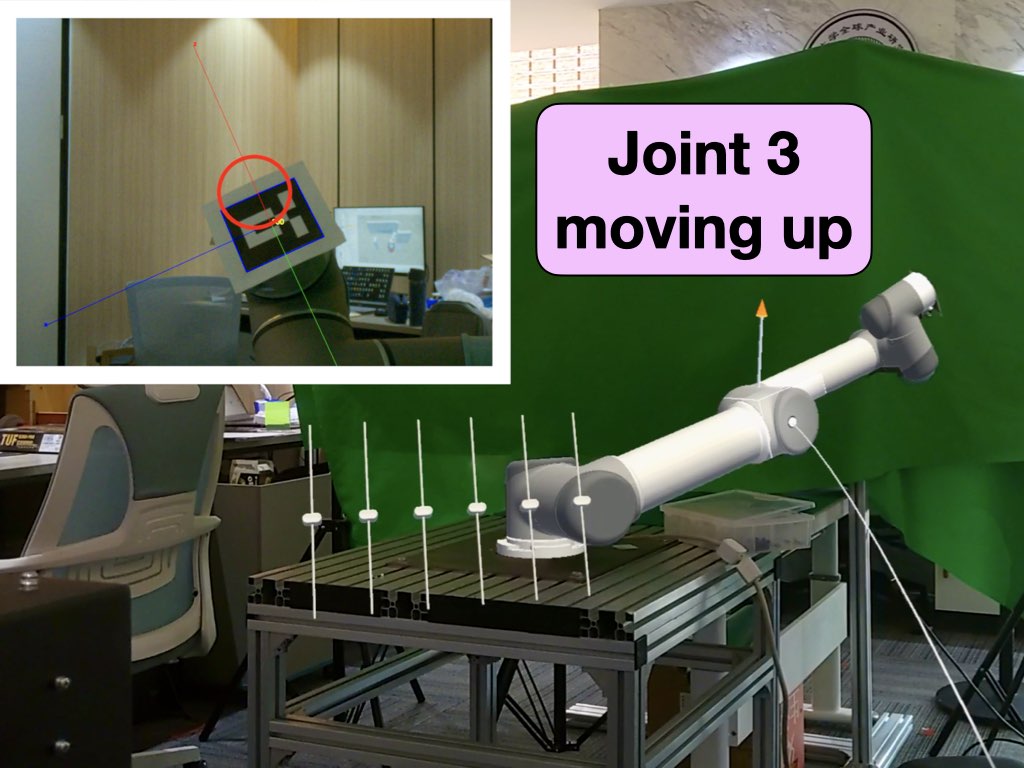} 
  } 
  \hspace{-0.4cm}
  \subfigure[]{ 
    \includegraphics[width=3.4cm]{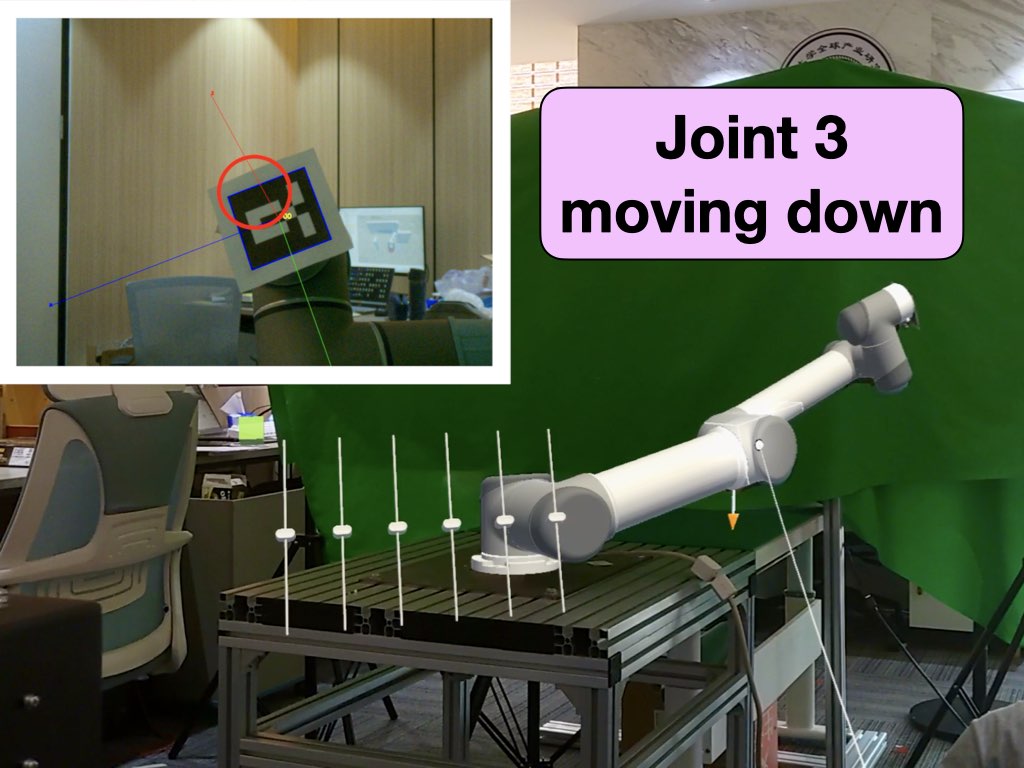} 
  } 
  \vspace{-3mm}
  \caption{Snapshots of task 2. The manipulator end-effector was performing a path-following task in vision space while the operator was guiding the manipulator body through the AR interface to avoid the collision.
    (a) $t = 0.0 \hspace{0.2mm} {\rm s}$: Initial state.
    (b) $t = 5.4 \hspace{0.2mm} {\rm s}$: The end-effector began to follow the circular path.
    (c) $t = 15.8 \hspace{0.2mm} {\rm s}$: A human subject placed a toolbox on the workbench, which raised the safety issue.
    (d) $t = 31.1 \hspace{0.2mm} {\rm s}$: The manipulator adjusted its pose according to the operator's inputs, while the end-effector's position in vision space remained unaffected.
    (e) $t = 57.3 \hspace{0.2mm} {\rm s}$: Guided by the operator, the manipulator resumed its original pose after the toolbox was removed.}
    \label{e2s}

\end{figure*}
\begin{figure*} [!tb]
  \centering 
  \subfigure[]{ 
    \includegraphics[height=2.4cm]{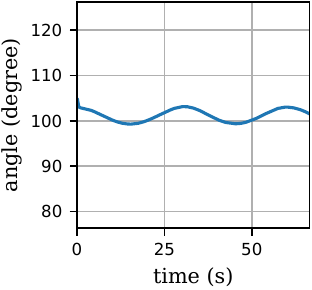}
  }
  \hspace{-0.35cm}
  \subfigure[]{ 
    \includegraphics[height=2.4cm]{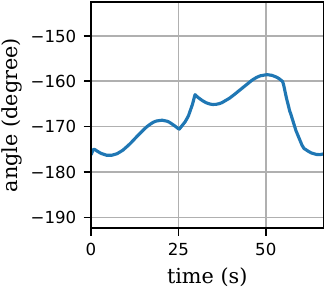}  
  } 
  \hspace{-0.35cm}
  \subfigure[]{
    \includegraphics[height=2.4cm]{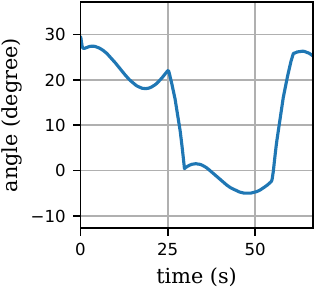}
  }
  \hspace{-0.35cm}
  \subfigure[]{ 
    \includegraphics[height=2.4cm]{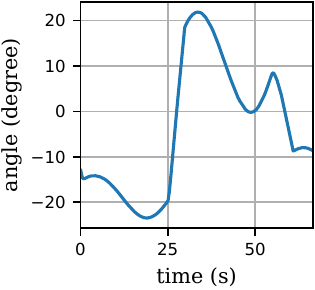} 
  } 
  \hspace{-0.35cm}
  \subfigure[]{ 
    \includegraphics[height=2.4cm]{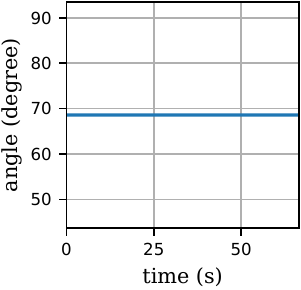} 
  } 
  \hspace{-0.35cm}
  \subfigure[]{ 
    \includegraphics[height=2.4cm]{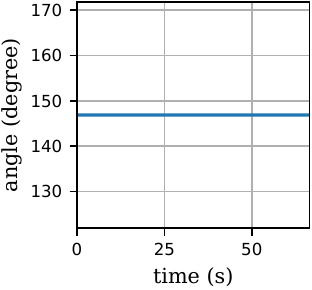} 
  }
  \vspace{-3mm}
  \caption{Joint motions in task 2. 
  (a) Joint 1; 
    (b) Joint 2;
    (c) Joint 3;
    (d) Joint 4;
    (e) Joint 5; 
    (f) Joint 6;}
    \label{e2angle}
    \vspace{-3mm}
\end{figure*}

\begin{figure}[!tb]
\centering 
  \subfigure[]{
    \includegraphics[height=3.7cm]{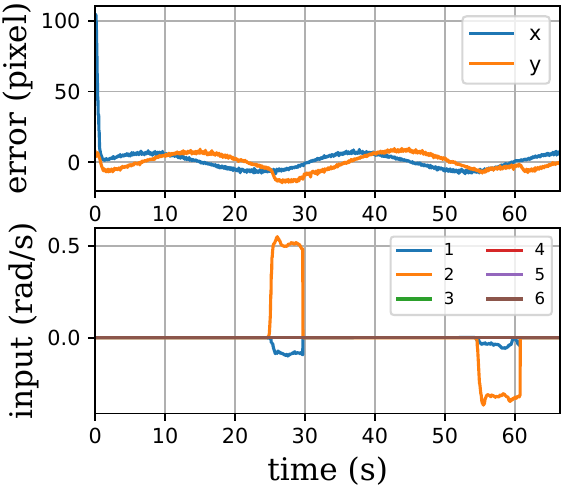}
  }
    \subfigure[]{ 
    \includegraphics[height=3.7cm]{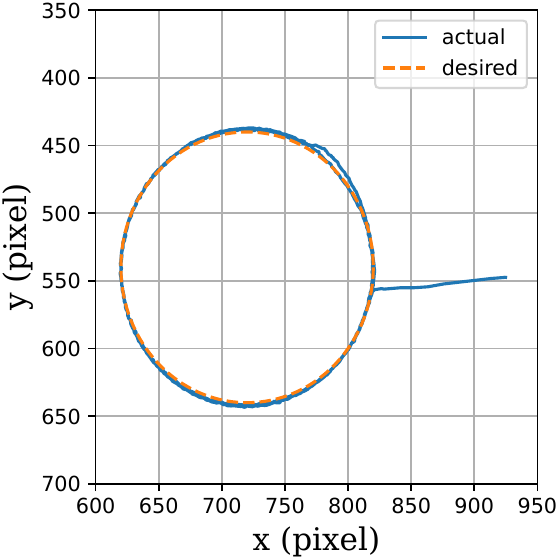} 
  } 
\vspace{-3mm}
  \caption{Results of task 2:
    (a) top: the position errors of the end-effector in vision space; bottom: the human intention applied to the joints
    (b) The trajectory of the end-effector in vision space. }
        \label{e2_pos_d}
    \vspace{-3mm}
\end{figure}
The control parameters in (\ref{xp}), (\ref{updateThetaz}), (\ref{updateThetak}), and (\ref{visualFeedback}) were set according to Table \ref{exp_para_table}, where $\bm I_2\in\Re^{2\times 2}$ is an identity matrix. The initial values of $\hat{\bm\theta}_z$ and $\hat{\bm\theta}_k$ were chosen randomly, and the control frequency was set as $30~{\rm Hz}$.
The experimental results are shown in Fig. \ref{e1_te}, in which the manipulator started to move to the desired position at $t=0.0~{\rm s}$. It was seen in Fig. \ref{e1_te}(a) that the position error in vision space was converged to zero in 1.6 seconds. 
Then the operator began to input the control efforts at $t\approx 6.0~{\rm s}$ (see Fig. \ref{e1_te}(a)), and the null-space control term was activated to smoothly react to it to drive the motion of redundant joints while maintaining the positioning task of the robot end-effector. The task error at steady state was less than 27 pixels, proving the realization of the positioning task.

An ablation study was also conducted to show the effectiveness of the adaptation of unknown parameters. A controller without adaptation was implemented by setting $\bm L_z$, $\bm L_k$ as zeros, where other parameters remained the same as Table I. As shown in Fig. \ref{adaptive_comp}(a), the convergence without adaptation (the upper one) is slower than the proposed controller (the lower one). Fig. \ref{adaptive_comp}(b) illustrates the estimated and actual depth in task 1, where the depth is updated and becomes closer to the actual value, and the adaptation stops when the position error reduces to zero.

In the second experiment, the manipulator was controlled to track a circular path in vision space. The moving velocity of the path was minimal, such that the moving target can be treated as a series of setpoints. During the task, a toolbox was placed around the base of the manipulator to simulate the presence of unforeseen changes. Such change may result in the collision, but it was outside the FOV of the camera and hence not detectable by the robot itself. The operator got involved in handling it by moving joint 3 away from the box. 
The whole process is shown in Fig. \ref{e2s}. In this task, the virtual robot manipulator was overlapped with the real manipulator, and hence the operator applied control efforts in an air tapping manner.

The desired path and the control parameters can be found in Table \ref{exp_para_table}.
As shown in Fig. \ref{e2_pos_d}, the position error of the end-effector was mainly induced from the periodic path-following in the main task and independent of human involvement. This was because the null-space interactive control term responded and isolated its influence to the main task. 
The motion of the joint angles is shown in Fig. \ref{e2angle}, where joint 2 to joint 4 moved to respond to the operator's command without affecting the task of the robot end-effector. 
The experimental results proved the feasibility and the effectiveness of the proposed control scheme in those scenarios where both the uncalibrated relationship and dynamic changes arose. The results can also be found in the video submission.

\section{Conclusions}
In this paper, an adaptive vision-based controller with null-space interaction has been proposed for redundant robots, which guarantees both the positioning accuracy in sensory space and compliance in null space.
Such a formulation allows humans to get involved in dealing with unforeseen changes without affecting the task of the robot end-effector. Online adaptation laws have been proposed to deal with unknown camera models, laying the foundation for robotic manipulation applications in uncalibrated environments.
The stability of the closed-loop system has been rigorously proved with Lyapunov methods, and the convergence of both the position error in vision space and damping model in null space is theoretically guaranteed. 
Experiments have been carried out to illustrate the performance of the proposed control scheme in different scenarios. Future works will be devoted to the extension to dynamics-based control tasks and validation on a 7-degrees-of-freedom force-control enabled industrial manipulator.
\newpage
\bibliographystyle{ieeetr}
\bibliography{ref}

\end{document}